\documentclass[a4paper]{article}

\usepackage{INTERSPEECH2022}
\usepackage[colorlinks=true, linkcolor=blue, citecolor=blue, urlcolor=blue]{hyperref}
\title{SHNU Multilingual Conversational Speech Recognition System for INTERSPEECH 2025 MLC-SLM Challenge \\ 
\thanks{$^{\star}$Yanhua Long is the corresponding author.}\vspace{-0.6cm}
\thanks{This work was sponsored by Natural Science Foundation of Shanghai (Grant No.25ZR1401277).}}

\name{Yuxiang Mei$^1$, Yuang Zheng$^1$,Dongxing Xu$^2$, Yanhua Long$^{1,2,\star}$}
\address{
  $^1$ Shanghai Normal University, Shanghai, China 
  \\$^2$Unisound AI Technology Co., Ltd., Beijing, China}
\email{153517myx@gmail.com, yanhua@shnu.edu.cn}

\begin{document}

\maketitle
\begin{abstract}
This paper describes SHNU multilingual conversational speech recognition system (SHNU-mASR, team name-``maybe"), submitted to Track 1 of the INTERSPEECH 2025 MLC-SLM Challenge.  Our system integrates a parallel-speech-encoder architecture with a large language model (LLM) to form a unified multilingual ASR framework. The parallel-speech-encoder consists of two pre-trained encoders, the Whisper-large-v3 encoder and mHuBERT-147 encoder. Their output embeddings are concatenated and fed into the LLM, enabling the model to leverage complementary acoustic and linguistic knowledge and achieve competitive performance. Moreover, we adopt a tri-stage training strategy to jointly update the low-rank adaptation modules and projector parameters of both the speech encoders and the LLM. In addition, we incorporate an additional language-aware prompt at the LLM input to enhance language-specific text generation.
The SHNU-mASR system achieves an overall character/word error rate (CER/WER) of 11.76\% on the blind evaluation set of the challenge, outperforming the official MLC-SLM baseline by 8.41 absolute CER/WER,  without increasing the baseline training data.
\end{abstract}
\vspace{-0.15cm}
\section{Introduction}

Automatic Speech Recognition (ASR) has advanced significantly, enabling accurate conversion of speech to text. Multilingual ASR (mASR) extends this capability to multiple languages within a single system, enhancing global communication and supporting low-resource languages. By learning shared representations, it improves both recognition accuracy and model efficiency, making it valuable for applications such as multilingual voice assistants and transcription services.

The INTERSPEECH 2025 MLC-SLM Challenge aims to bridge the gap by encouraging the development of multilingual language models for conversational speech (MLC-SLM). The challenge consists of two tasks:  (1) multilingual conversational speech recognition, and (2) multilingual conversational speech diarization and recognition. This paper focuses on Task 1,  aiming to improve recognition accuracy in multilingual conversational scenarios.

In recent years, many multilingual ASR systems were proposed, such as Whisper~\cite{radford2023robust}, a simple end-to-end approach, implemented as an encoder-decoder Transformer\cite{vaswani2017attention} and trained in a multitask manner and large diverse dataset. Many other works have also been proposed to enhance Whisper, such as LoRA-Whisper~\cite{song2024lora}, Multilingual DistilWhisper~\cite{ferraz2024multilingual}, and WhisperX ~\cite{bain2023whisperx}, which improves alignment accuracy for word-level timestamps and long-form audio. Distil-Whisper~\cite{gandhi2023distil}, a smaller, faster version of Whisper distilled with pseudo labels, retaining robustness and enabling efficient speculative decoding.
In ~\cite{hu2023mixture,wang2023language,ma2025blr,ye2024sc,zhao2024saml}, Mixture-of-Expert(MoE) based methods, leveraging expert specialization and sparse activation to improve performance in multilingual ASR, scalability, and efficiency across diverse languages and streaming scenarios;
Many well pre-trained models with self-supervised learning (SSL) have also been widely used for multilingual ASR, such as in mHuBERT-147~\cite{boito2024mhubert}, a compact HuBERT model trained on 90K hours of open data with strong ASR and LID performance, MSR-86K~\cite{li2024msr},which enables SSL training through a large-scale open-source multilingual corpus of 86K hours, ML-SUPERB2.0~\cite{shi2024ml} a benchmark specifically designed to evaluate the generalization and adaptability of SSL-based speech models across languages and tasks.
 In addition, multilingual datasets such as Multilingual LibriSpeech (MLS)~\cite{pratap2020mls}, MSR-86K~\cite{li2024msr}, and industrial-scale corpora(e.g., Anatomy of Industrial-Scale Multilingual ASR~\cite{ramirez2024anatomy}) play a crucial role in advancing the field. These previous works have significantly improved multilingual ASR performance. However, large language models (LLMs)~\cite{touvron2023llama,devlin2019bert} have recently demonstrated remarkable capabilities in language understanding and generation. Recent efforts such as Ideal-LLM~\cite{xue2024ideal}, SALMONN~\cite{tang2023salmonn}, MinMo~\cite{chen2025minmo}, and Qwen2-Audio~\cite{chu2024qwen2} explore the integration of speech encoders with LLMs to enable multilingual ASR, speech translation, and universal audio-language understanding. How to effectively integrate LLMs into mASR systems remains an important and promising direction for further research.

Inspired by the MLC-SLM challenge, this paper explores new architecture and tri-stage system training 
methods to integrate a  LLM with a parallel-speech-encoder, aiming to boost the multilingual ASR for 
real-world conversational speech. Specifically, we employ a combination  of Low-Rank Adaptation (LoRA)~\cite{hu2022lora} and full fine-tuning to 
update different modules within the parallel-speech-encoder. A projector module is then introduced  to 
connect the aoucstic representations with the LoRA-tuned LLM, further enhanced by a language-aware prompt 
as the input stage.  Our SHNU-mASR achieves an overall CER/WER of 14.37\% on the development set. On the blind evaulation set of the INTERSPEECH MLC-SLC Challenge,
it achieves a CER/WER of 11.76\%, which is a very competitive performance in track 1 with only 1500hrs training data. 
\vspace{-0.15cm}
\section{Proposed SHNU-mASR  System}

Our proposed SHNU-mASR system mainly composed of three modules:  a parallel-speech-encoder, 
a projector and a language-aware prompted LLM. The whole architecture of SHNU-mASR is illustrated 
in Fig.~\ref{fig:enter-label}. 

\subsection{Parallel-speech-encoder}
The parallel speech encoder consists of two complementary encoders: Whisper~\cite{radford2023robust} and mHuBERT~\cite{boito2024mhubert}. These encoders are used in parallel to extract rich speech representations from the input waveform. While Whisper is a supervised encoder trained on large-scale multilingual speech-text pairs, mHuBERT is a self-supervised model pretrained to capture universal acoustic features across languages. By combining their outputs, the system benefits from both high-level semantic information and robust acoustic generalization, making it well-suited for multilingual ASR tasks.

To adapt these encoders efficiently for the MLC-SLM task, we adopt different parameter update strategies based on model size and training cost:

\begin{itemize}
\item For the Whisper-large model (over 1.5B parameters), full fine-tuning is computationally intensive and memory-demanding. To address this, we apply Low-Rank Adaptation (LoRA)~\cite{hu2022lora}, which introduces trainable low-rank matrices into selected layers while keeping the original weights frozen, enabling efficient and lightweight adaptation.
\item In contrast, mHuBERT is smaller and more computationally feasible. Owing to its self-supervised pretraining and smaller capacity, it exhibits greater adaptability to domain-specific speech, as observed by Geng et al.~\cite{geng2024unveiling}. Therefore, we apply full fine-tuning to mHuBERT to maximize its ability to model language- and task-specific features.
\end{itemize}

\subsubsection{Parameter-efficient Adaptation with LoRA}
To reduce training cost for the large-scale Whisper encoder, we adopt LoRA~\cite{hu2022lora} to fine-tune the model efficiently. LoRA modules are injected into the self-attention layers, specifically modifying the query and value projection matrices $W_q$ and $W_v$, while keeping the rest of the encoder parameters frozen.

Following the standard LoRA formulation~\cite{hu2022lora}, each modified matrix is decomposed into a low-rank form as:
\begin{equation}
W_q^{\text{LoRA}} = W_q + \Delta W_q = W_q + \alpha A_q B_q,
\end{equation}
\begin{equation}
W_v^{\text{LoRA}} = W_v + \Delta W_v = W_v + \alpha A_v B_v,
\end{equation}
where $A_q, A_v \in \mathbb{R}^{d \times r}$ and $B_q, B_v \in \mathbb{R}^{r \times d}$ are the learnable low-rank matrices, $r$ is the LoRA rank, and $\alpha$ is a scaling factor. This structure enables efficient adaptation by only updating a small number of additional parameters.

Given an input acoustic feature sequence $x = \{x_1, x_2, \ldots, x_T\}$, the Whisper encoder with LoRA modules computes the output hidden states as:
\begin{equation}
h^{\text{Whisper}} = \text{Encoder}_{\theta + \Delta\theta}(x)
\end{equation}
where $\theta$ denotes the frozen original parameters and $\Delta\theta$ represents the set of trainable LoRA parameters.

\subsubsection{Full Fine-tuning of mHuBERT}
During training, all parameters of the mHuBERT encoder are updated, allowing the model to learn fine-grained acoustic and phonetic variations present in conversational multilingual speech. Given the same input $x = \{x_1, x_2, \ldots, x_T\}$, the output hidden states from mHuBERT are computed as:
\begin{equation}
h^{\text{mHuBERT}} = \text{Encoder}_{\theta}(x)
\end{equation}
where parameters $\theta$ are trainable and updated during training.

\begin{figure}[htbp]
    \centering
    \includegraphics[width=1.0\linewidth]{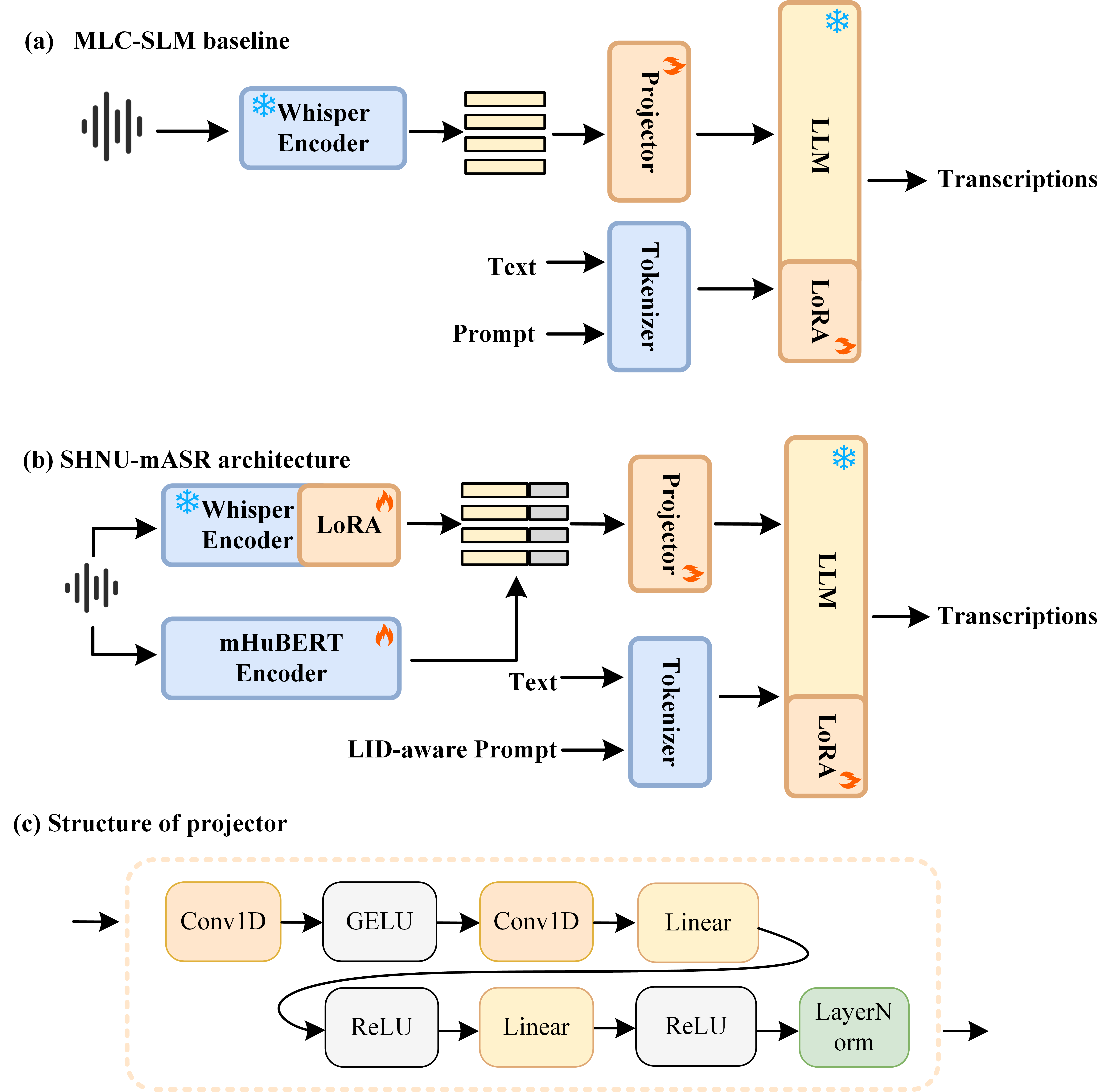}
    \caption{Overall model structure. (a)The baseline structure of MLC-SLM. (b)SHNU-mASR adopts a parallel-speech-encoder design, followed by a projection module before integration with the LLM.(c) Projector block architecture details.}
    \label{fig:enter-label}
\end{figure}
\vspace{-0.6cm}
\subsection{Projector}

To leverage the complementary strengths of Whisper and mHuBERT, we concatenate their output hidden states to form a unified representation for the downstream large language model (LLM). Specifically, the hidden states $h^{\text{Whisper}}$ and $h^{\text{mHuBERT}}$ are concatenated along the feature dimension:

\begin{equation}
h^{\text{concat}} = \text{Concat}(h^{\text{Whisper}},\ h^{\text{mHuBERT}}) \in \mathbb{R}^{T \times (d_1 + d_2)}
\end{equation}
where $d_1$ and $d_2$ are the output dimensions of the Whisper and mHuBERT encoders, respectively, and $T$ is the sequence length. The fused representation $h^{\text{concat}}$ is then passed to a projection module and subsequently fed into the LLM.

As illustrated in Figure~\ref{fig:enter-label} (c), the projector first applies a 1D convolution with kernel size 3 to capture local temporal features. This is followed by a strided convolution to downsample the sequence length, reducing computational cost. The downsampled features are then passed through a two-layer feed-forward network (MLP) to project the representation into the target LLM input dimension. Finally, LayerNorm is applied to stabilize the feature distribution before passing it to the LLM.
The final projected representation is denoted as:

\begin{equation}
E_{\text{s}} = \text{Projector}(h^{\text{concat}}) \in \mathbb{R}^{T' \times d'}
\end{equation}
where $T'$ is the downsampled sequence length, and $d'$ is the target feature dimension required by the LLM.
This design enables the LLM to access both high-level semantic features from Whisper and robust acoustic cues from mHuBERT, enhancing its capacity to model multilingual conversational speech.

\subsection{Language-aware LLM}

To guide the LLM toward language-aware transcription, we introduce a language-specific prompt prepended to each input. This prompt is a natural language instruction indicating the target transcription language, such as:

\begin{quote}
\texttt{Please transcribe the following audio in \{lang\}:}
\end{quote}

Given a batch of inputs with corresponding target languages and target transcriptions, the prompts and target texts are tokenized and embedded using the LLM’s tokenizer and embedding layer:

\begin{equation}
E_p = \text{Embedding}(\text{Tokenizer}(P))
\end{equation}

\begin{equation}
E_t = \text{Embedding}(\text{Tokenizer}(T_{rans}))
\end{equation}
where \( P \) denotes the language-aware prompt, and \(T_{rans} \) denotes the target transcription tokens. Finally, the full input to the LLM is the concatenation of these three components along the time dimension:

\begin{equation}
LLM_{input} = \text{Concat}(E_p,\ E_s,\ E_t)
\end{equation}

This design allows the LLM to jointly consider language intent (\( E_p \)), acoustic features (\( E_s \)), and supervision signals (\( E_t \)) during training or inference.

\subsection{Tri-stage Model Training}

We adopt a tri-stage training strategy to progressively align speech and language representations while ensuring stable convergence for our parallel-speech-encoder ASR model. Initially, we attempted to unfreeze the speech encoders, the projector, and the LLM LoRA~\cite{hu2022lora} modules simultaneously. However, this approach failed to converge and resulted in poor performance, consistent with observations in previous work~\cite{geng2024unveiling}. To address this, we propose the following three-stage training process:

\begin{itemize}
  \item \textbf{Stage 1: Projector pretraining.} We freeze all other components and train only the projector to align the fused hidden states from Whisper and mHuBERT into a representation suitable for the LLM.

  \item \textbf{Stage 2: Encoder adaptation.} We unfreeze the LoRA modules of the Whisper encoder for parameter-efficient fine-tuning, and fully unfreeze the mHuBERT encoder for full fine-tuning, while keeping the LLM frozen. The projector remains trainable during this stage. This enables the encoders to adapt to the target task while balancing computational cost.
  \item \textbf{Stage 3: LLM adaptation.} We unfreeze the LoRA modules of the LLM and jointly fine-tune them together with the projector and both encoders (LoRA for Whisper, full fine-tuning for mHuBERT). This stage allows the LLM to adapt to the fused speech representations and improve transcription quality in a multilingual prompt-based setting.
\end{itemize}

This staged training strategy stabilizes optimization and prevents the training collapse observed when all components are updated simultaneously.

\section{Experment Setup}
\subsection{Datasets}

The training set consists of multilingual conversational speech data across 11 languages: English (en), French (fr), German (de), Italian (it), Portuguese (pt), Spanish (es), Japanese (ja), Korean (ko), Russian (ru), Thai (th), and Vietnamese (vi).
Each audio clip features a natural dialogue between two speakers discussing randomly assigned topics. Recordings were captured in quiet indoor environments using devices such as iPhones, and are annotated with accurate timestamps and speaker labels to support both ASR and speaker diarization tasks.
The English subset includes approximately 500 hours of speech from various regions (e.g., US, UK, Australia, India, and the Philippines), while each of the other languages contributes around 100 hours, yielding a total of roughly 1500 hours of multilingual data.
The development set (Dev) has the same distribution as the training set, containing about 2 hours per language. It is officially used for evaluation.

In contrast to the baseline system, which directly uses the official Dev set for testing, our training pipeline avoids this overlap to better evaluate model generalization. Specifically, we reserve the official Dev set exclusively for testing. During training, we construct an internal validation set by randomly selecting 2 hours per language from the training data, which is used for model supervision and early stopping. This design ensures a clear separation between training and evaluation data.
\subsection{Model Configurations}

We employ mHuBERT-147~\cite{boito2024mhubert} and Whisper Large-v3~\cite{radford2023robust} encoder as our parallel-speech-encoder. The outputs of these two encoders are concatenated to form a 2048-dimensional joint representation, which is then passed through a projector module. This projector performs a 4 times downsampling along the time dimension and maps the 2048-dimensional input to a 3584-dimensional vector, matching the input dimensionality required by the LLM, which is Qwen2.5-7B~\cite{yang2025qwen3}.

Unlike full fine-tuning, LoRA~\cite{hu2022lora} allows efficient adaptation by introducing low-rank trainable matrices while keeping the original model weights frozen. We configure LoRA with alpha = 8, rank = 16 for LLM and alpha =16 ,rank =8 for Whisper. The training process is conducted using four NVIDIA A100 80G GPUs with BF16 precision. We use the Adam optimizer with a learning rate of 1.0e-04, beta = (0.9, 0.99), epsilon = 1.0e-06, and weight decay = 0.01. A warm-up strategy with 5000 steps is used, followed by inverse square root decay. Gradient clipping is applied with a maximum norm of 5 to prevent gradient explosion. We use dynamic batching based on the total number of audio frames, with a maximum of 120 seconds of audio per batch. All models are trained for 6 epochs.
\subsection{Evaluation Metrics}

Following standard multilingual ASR evaluation protocols, we compute Character Error Rate (CER) for languages with non-Latin scripts, specifically Japanese, Korean, and Thai. For all other languages, Word Error Rate (WER) is used as the evaluation metric. All error rates are calculated using the meeteval toolkit~\cite{von2023meeteval}, which provides consistent and reliable scoring across different languages and segmentation conditions.

\section{Results and Discussions}
\subsection{Baseline with Inference-Time Text Normalization and LID-Aware Prompting on the MLC-SLM dev-set}

\begin{table}[h]
\centering

\caption{Average CER/WER (\%) on the MLC-SLM dev-set for different models and inference-time settings.}
\label{tab:baseline_avg}
\begin{tabular}{lc}
\toprule
\textbf{Model} & \textbf{CER/WER} \\
\midrule
Vanilla Whisper-large-v3 & 16.82 \\
Baseline-Qwen            & 21.49 \\
Baseline-Qwen + Norm     & 18.75 \\
Baseline-Qwen + Norm + LID & 18.03 \\
\bottomrule
\end{tabular}
\end{table}
 
The inference-time text normalization (Norm) process removes repetitive tokens from the LLM output during inference, which helps to reduce spurious repetitions and improves text quality. The LID-aware prompting guides the model by explicitly indicating the language of the input, helping to disambiguate multilingual content and thus enhancing recognition performance.

Table~\ref{tab:baseline_avg} shows that the baseline Qwen model initially underperforms compared to the vanilla Whisper-large-v3. Applying output deduplication during inference significantly improves performance by mitigating errors caused by repeated tokens. Further incorporating LID-aware prompting only during inference time yields additional performance gains, as the model benefits from explicit language cues that reduce confusion across languages. These results confirm the effectiveness of inference-time techniques in enhancing model robustness without modifying the underlying model parameters.

\subsection{Ablation results with different parallel-speech-encoder structure}

\begin{table}[h]
\centering

\caption{Impact of different parallel-speech-encoder structures using only 1/4 of the training data. Average CER/WER (\%) is reported on the MLC-SLM dev-set.}
\label{tab:ablation_encoder}
\begin{tabular}{lc}
\toprule
\textbf{Model} & \textbf{CER/WER} \\
\midrule
Baseline-Qwen                          & 21.13 \\
Whisper + Qwen2Audio encoder           & 18.63 \\
Whisper + mHuBERT encoder              & 19.32 \\
\bottomrule
\end{tabular}
\end{table}

Table~\ref{tab:ablation_encoder} shows the comparison of different parallel-speech-encoder structures under a low-resource setting where only one-quarter of the baseline training data is used. Each row represents a different model configuration:
\begin{itemize}
    \item \textbf{Baseline-Qwen}: The challenge official baseline configuration with only Whisper speech encoder.
    \item \textbf{Whisper + Qwen2Audio encoder}: A parallel structure where Whisper and a Qwen2Audio-based speech encoder process input speech simultaneously.
    \item \textbf{Whisper + mHuBERT encoder}: A parallel structure combining Whisper with a pre-trained mHuBERT encoder.
\end{itemize}

In all configurations involving additional encoders, we freeze the speech encoder parameters and only fine-tune the projector module (used to fuse features from the dual encoders) and the LLM using LoRA fine-tuning. This ablation aims to evaluate the effectiveness of introducing an auxiliary speech encoder without increasing the number of trainable parameters in the encoder backbone.

As shown in Table~\ref{tab:ablation_encoder}, both parallel-speech-encoder setups outperform the baseline. Adding the Qwen2-Audio~\cite{chu2024qwen2} encoder brings the largest gain, reducing the CER/WER from 21.13\% to 18.63\%. The mHuBERT-enhanced version also shows improvement (19.32\%), although slightly less than Qwen2-Audio. These results suggest that a parallel-speech-encoder structure can leverage complementary representations to improve recognition, even when the added encoder is frozen during training.

Despite Qwen2-Audio achieving slightly better results in this low-resource setting, we ultimately adopt the \textbf{Whisper + mHuBERT} structure in our final SHNU-mASR system. This choice is motivated by the better generalization performance observed in full-data experiments and the robustness of mHuBERT across diverse language conditions.

\subsection{Results with 1500hrs training data}

\begin{table}[h]
\centering
\caption{Average CER/WER (\%) on the MLC-SLM challenge dev-set and blind eval-set using the full 1500-hour training data.}
\label{tab:final_results}
\begin{tabular}{lcc}
\toprule
\textbf{Model} & \textbf{Dev-set} & \textbf{Eval-set} \\
\midrule
Vanilla Whisper large-v3       & 16.82 & --    \\
Baseline-Qwen + Norm + LID & 18.03 & 20.17 \\
SHNU-mASR              & \textbf{14.37} & \textbf{11.76} \\
\bottomrule
\end{tabular}
\end{table}
Table~\ref{tab:final_results} presents average CER/WER on both the MLC-SLM development set (dev-set) and the blind evaluation set (eval-set). The \textbf{Vanilla Whisper large-v3} and the \textbf{Baseline-Qwen + LID + Norm} are the same as shown in Table~\ref{tab:baseline_avg}. \textbf{SHNU-mASR} is our proposed system using Whisper with an auxiliary mHuBERT encoder, a projection layer, and Qwen with LoRA fine-tuning with LID-aware prompt.

Compared to the Baseline-Qwen system, SHNU-mASR reduces the average error rate from 18.03\% to 14.37\% on the dev-set (a 3.66 absolute point gain). More significantly, on the blind eval-set, the error rate drops from 20.17\% to 11.76\%, which represents a \textbf{41.7\% relative reduction} in errors.These results confirm that SHNU-mASR generalizes better across domains and accents. The performance gain is attributed to:
\begin{itemize}
    \item The integration of an additional speech encoder (mHuBERT) which enhances acoustic feature diversity.
    \item Fine-tuning only the projector and LLM via LoRA, which prevents overfitting and supports rapid adaptation.
    \item The modular design enabling better synergy between the speech encoders and the language model.
\end{itemize}

SHNU-mASR significantly outperforms both the vanilla Whisper and the Baseline-Qwen system. The consistent improvements across dev and eval sets demonstrate the robustness and effectiveness of our architecture in multilingual ASR.

\section{Conclusions }

This paper introduces our submission to the INTERSPEECH 2025 MLC-SLM Challenge-Multilingual Conversational Speech Recognition track. Our SHNU-mASR system integrates a parallel-speech-encoder architecture with a LoRA-tuned large language model, enhanced by a tri-stage training strategy and language-specific prompting. Without increasing the baseline training data, our system achieves competitive multilingual ASR performance and a 41.7\% relative reduction on the MLC-SLM blind evaluation set.
\bibliographystyle{IEEEtran}

\bibliography{IS2022_paper_kit/LaTeX/IEEEtran/bibtex/MlC-SLM}

\begin{thebibliography}{10}
\providecommand{\url}[1]{#1}
\csname url@samestyle\endcsname
\providecommand{\newblock}{\relax}
\providecommand{\bibinfo}[2]{#2}
\providecommand{\BIBentrySTDinterwordspacing}{\spaceskip=0pt\relax}
\providecommand{\BIBentryALTinterwordstretchfactor}{4}
\providecommand{\BIBentryALTinterwordspacing}{\spaceskip=\fontdimen2\font plus
\BIBentryALTinterwordstretchfactor\fontdimen3\font minus \fontdimen4\font\relax}
\providecommand{\BIBforeignlanguage}[2]{{%
\expandafter\ifx\csname l@#1\endcsname\relax
\typeout{** WARNING: IEEEtran.bst: No hyphenation pattern has been}%
\typeout{** loaded for the language `#1'. Using the pattern for}%
\typeout{** the default language instead.}%
\else
\language=\csname l@#1\endcsname
\fi
#2}}
\providecommand{\BIBdecl}{\relax}
\BIBdecl

\bibitem{radford2023robust}
A.~Radford, J.~W. Kim, T.~Xu, G.~Brockman, C.~McLeavey, and I.~Sutskever, ``Robust speech recognition via large-scale weak supervision,'' in \emph{Proceedings of International Conference on Machine Learning (ICML)}, 2023, pp. 28\,492--28\,518.

\bibitem{vaswani2017attention}
A.~Vaswani, N.~Shazeer, N.~Parmar, J.~Uszkoreit, L.~Jones, A.~N. Gomez, {\L}.~Kaiser, and I.~Polosukhin, ``{Attention Is All You Need},'' \emph{Advances in neural information processing systems}, vol.~30, 2017.

\bibitem{song2024lora}
Z.~Song, J.~Zhuo, Y.~Yang, Z.~Ma, S.~Zhang, and X.~Chen, ``Lora-{W}hisper: Parameter-{E}fficient and {E}xtensible {M}ultilingual {ASR},'' in \emph{Proceedings of Interspeech}, 2024, pp. 3934--3938.

\bibitem{ferraz2024multilingual}
T.~P. Ferraz, M.~Z. Boito, C.~Brun, and V.~Nikoulina, ``Multilingual distilwhisper: Efficient distillation of multi-task speech models via language-specific experts,'' in \emph{Proceedings of IEEE International Conference on Acoustics, Speech and Signal Processing (ICASSP)}, 2024, pp. 10\,716--10\,720.

\bibitem{bain2023whisperx}
M.~Bain, J.~Huh, T.~Han, and A.~Zisserman, ``Whisperx: Time-accurate speech transcription of long-form audio,'' in \emph{Proceedings of Interspeech}, 2023, pp. 4489--4493.

\bibitem{gandhi2023distil}
S.~Gandhi, P.~von Platen, and A.~M. Rush, ``Distil-whisper: Robust knowledge distillation via large-scale pseudo labelling,'' \emph{arXiv preprint arXiv:2311.00430}, 2023.

\bibitem{hu2023mixture}
K.~Hu, B.~Li, T.~Sainath, Y.~Zhang, and F.~Beaufays, ``Mixture-of-{E}xpert {C}onformer for {S}treaming {M}ultilingual {ASR},'' in \emph{Proceedings of Interspeech}, 2023, pp. 3327--3331.

\bibitem{wang2023language}
W.~Wang, G.~Ma, Y.~Li, and B.~Du, ``Language-routing mixture of experts for multilingual and code-switching speech recognition,'' in \emph{Proceedings of Interspeech}, 2023, pp. 1389--1393.

\bibitem{ma2025blr}
G.~Ma, W.~Wang, L.~Zhou, Y.~Yang, Y.~Li, and B.~Du, ``{BLR-M}o{E}: Boosted language-routing mixture of experts for domain-robust multilingual {E2E ASR},'' in \emph{Proceedings of IEEE International Conference on Acoustics, Speech and Signal Processing (ICASSP)}, 2025.

\bibitem{ye2024sc}
S.~Ye, S.~Chen, X.~Hu, and X.~Xu, ``{SC-M}o{E}: Switch conformer mixture of experts for unified streaming and non-streaming code-switching {ASR},'' in \emph{Proceedings of Interspeech}, 2024, pp. 3999--4003.

\bibitem{zhao2024saml}
Q.~Zhao, G.~Sun, C.~Zhang, M.~Xu, and T.~F. Zheng, ``Saml: Speaker adaptive mixture of lora experts for end-to-end {ASR},'' in \emph{Proceedings of Interspeech}, 2024, pp. 777--781.

\bibitem{boito2024mhubert}
M.~{Zanon Boito}, V.~Iyer, N.~Lagos, L.~Besacier, and I.~Calapodescu, ``m{H}u{BERT}-147: A compact multilingual {H}u{BERT} model,'' in \emph{Proceedings of Interspeech}, 2024, pp. 3939--3943.

\bibitem{li2024msr}
S.~Li, Y.~You, X.~Wang, Z.~Tian, K.~Ding, and G.~Wan, ``{MSR-86K}: An evolving, multilingual corpus with 86,300 hours of transcribed audio for speech recognition research,'' in \emph{Proceedings of Interspeech}, 2024, pp. 1245--1249.

\bibitem{shi2024ml}
J.~Shi, S.-H. Wang, W.~Chen, M.~Bartelds, V.~{Bannihatti Kumar}, J.~Tian, X.~Chang, D.~Jurafsky, K.~Livescu, H.~yi~Lee, and S.~Watanabe, ``{ML-SUPERB} 2.0: Benchmarking multilingual speech models across modeling constraints, languages, and datasets,'' in \emph{Proceedings of Interspeech}, 2024, pp. 1230--1234.

\bibitem{pratap2020mls}
V.~Pratap, Q.~Xu, A.~Sriram, G.~Synnaeve, and R.~Collobert, ``{MLS}: A large-scale multilingual dataset for speech research,'' in \emph{Proceedings of Interspeech}, 2020, pp. 2757--2761.

\bibitem{ramirez2024anatomy}
F.~M. Ramirez, L.~Chkhetiani, A.~Ehrenberg, R.~McHardy, R.~Botros, Y.~Khare, A.~Vanzo, T.~Peyash, G.~Oexle, M.~Liang, I.~Sklyar, E.~Fakhan, A.~Etefy, D.~McCrystal, S.~Flamini, D.~Donato, and T.~Yoshioka, ``Anatomy of industrial scale multilingual {ASR},'' \emph{CoRR}, vol. abs/2404.09841, 2024.

\bibitem{touvron2023llama}
H.~Touvron, T.~Lavril, G.~Izacard, X.~Martinet, M.~Lachaux, T.~Lacroix, B.~Rozi{\`{e}}re, N.~Goyal, E.~Hambro, F.~Azhar, A.~Rodriguez, A.~Joulin, E.~Grave, and G.~Lample, ``{LLaMA}: Open and efficient foundation language models,'' \emph{CoRR}, vol. abs/2302.13971, 2023.

\bibitem{devlin2019bert}
J.~Devlin, M.-W. Chang, K.~Lee, and K.~Toutanova, ``{BERT}: Pre-training of deep bidirectional transformers for language understanding,'' in \emph{Proceedings of North American Chapter of the Association for Computational Linguistics}, 2019.

\bibitem{xue2024ideal}
H.~Xue, W.~Ren, X.~Geng, K.~Wei, L.~Li, Q.~Shao, L.~Yang, K.~Diao, and L.~Xie, ``Ideal-{LLM}: Integrating dual encoders and language-adapted {LLM} for multilingual speech-to-text,'' \emph{arXiv preprint arXiv:2409.11214}, 2024.

\bibitem{tang2023salmonn}
C.~Tang, W.~Yu, G.~Sun, X.~Chen, T.~Tan, W.~Li, L.~Lu, Z.~MA, and C.~Zhang, ``{SALMONN}: Towards generic hearing abilities for large language models,'' in \emph{Proceedings of The Twelfth International Conference on Learning Representations}, 2024.

\bibitem{chen2025minmo}
Q.~Chen, Y.~Chen, Y.~Chen, M.~Chen, Y.~Chen, C.~Deng, Z.~Du, R.~Gao, C.~Gao, Z.~Gao \emph{et~al.}, ``Min{M}o: A multimodal large language model for seamless voice interaction,'' \emph{arXiv preprint arXiv:2501.06282}, 2025.

\bibitem{chu2024qwen2}
Y.~Chu, J.~Xu, Q.~Yang, H.~Wei, X.~Wei, Z.~Guo, Y.~Leng, Y.~Lv, J.~He, J.~Lin \emph{et~al.}, ``Qwen2-audio technical report,'' \emph{arXiv preprint arXiv:2407.10759}, 2024.

\bibitem{hu2022lora}
E.~J. Hu, Y.~Shen, P.~Wallis, Z.~Allen-Zhu, Y.~Li, S.~Wang, L.~Wang, W.~Chen \emph{et~al.}, ``{LoRA}: Low-rank adaptation of large language models.'' \emph{Proceedings of ICLR}, vol.~1, no.~2, p.~3, 2022.

\bibitem{geng2024unveiling}
X.~Geng, T.~Xu, K.~Wei, B.~Mu, H.~Xue, H.~Wang, Y.~Li, P.~Guo, Y.~Dai, L.~Li \emph{et~al.}, ``Unveiling the potential of {LLM}-based asr on chinese open-source datasets,'' in \emph{Proceedings of International IEEE 14th Symposium on Chinese Spoken Language Processing (ISCSLP)}, 2024, pp. 26--30.

\bibitem{yang2025qwen3}
A.~Yang, A.~Li, B.~Yang, B.~Zhang, B.~Hui, B.~Zheng, B.~Yu, C.~Gao, C.~Huang, C.~Lv \emph{et~al.}, ``Qwen3 technical report,'' \emph{arXiv preprint arXiv:2505.09388}, 2025.

\bibitem{von2023meeteval}
T.~von Neumann, C.~Boeddeker, M.~Delcroix, and R.~Haeb-Umbach, ``{MeetEval}: A toolkit for computation of word error rates for meeting transcription systems,'' in \emph{Proceedings of 7th International Workshop on Speech Processing in Everyday Environments (CHiME 2023)}, 2023, pp. 27--32.

\end{thebibliography}


\end{document}